\documentclass[10pt,twocolumn,letterpaper]{article}

\usepackage{wacv}              

\usepackage{graphicx}
\usepackage{amsmath}
\usepackage{amssymb}
\usepackage{booktabs}
\usepackage{comment}
\usepackage{makecell}

\usepackage[pagebackref,breaklinks,colorlinks]{hyperref}

\usepackage[hyphens]{url}
\usepackage{colortbl}
\usepackage[dvipsnames]{xcolor}
\usepackage[capitalize]{cleveref}
\crefname{section}{Sec.}{Secs.}
\Crefname{section}{Section}{Sections}
\Crefname{table}{Table}{Tables}
\crefname{table}{Tab.}{Tabs.}

\def\wacvPaperID{1649} 
\def\confName{WACV}
\def\confYear{2025}

\usepackage{multirow}

\begin{document}

\title{UAL-Bench: The First Comprehensive Unusual Activity Localization Benchmark}

\author{Hasnat Md Abdullah\\
{\tt\small Texas A\&M University} \\
{\tt\small hasnat.md.abdullah@tamu.edu}
\and
Tian Liu\\
{\tt\small Texas A\&M University} \\
{\tt\small tian.liu@tamu.edu}
\and
Kangda Wei\\
{\tt\small Texas A\&M University} \\
{\tt\small kangda@tamu.edu}
\and
Shu Kong\\
{\tt\small Texas A\&M University} \\
{\tt\small University of Macau} \\
{\tt\small shu@tamu.edu, skong@um.edu.mo}
\and
Ruihong Huang\\
{\tt\small Texas A\&M University} \\
{\tt\small huangrh@tamu.edu}
}

\maketitle

\begin{abstract}
\noindent 
Localizing unusual activities, such as human errors or surveillance incidents, in videos holds practical significance. 
However, current video understanding models struggle with localizing these unusual events likely because of their insufficient representation in models' pretraining datasets. 
To explore foundation models' capability in localizing unusual activity, 
we introduce UAL-Bench, a comprehensive benchmark for unusual activity localization, featuring three video datasets: UAG-OOPS, UAG-SSBD, UAG-FunQA, and an instruction-tune dataset: OOPS-UAG-Instruct, to improve model capabilities. UAL-Bench evaluates three approaches: Video-Language Models (Vid-LLMs), instruction-tuned Vid-LLMs, and a novel integration of Vision-Language Models and Large Language Models (VLM-LLM). Our results show the VLM-LLM approach excels in localizing short-span unusual events and predicting their onset (start time) more accurately than Vid-LLMs. We also propose a new metric, $R@1, TD\leq p$, to address limitations in existing evaluation methods. Our findings highlight the challenges posed by long-duration videos, particularly in autism diagnosis scenarios, and the need for further advancements in localization techniques. Our work not only provides a benchmark for unusual activity localization but also outlines the key challenges for existing foundation models, 
suggesting future research directions on this important task.
   
\end{abstract}


\vspace{-0.5cm}
\section{Introduction}
\begin{figure}[t]
  \includegraphics[width=\columnwidth, trim={0cm 3cm 0cm 0cm}]{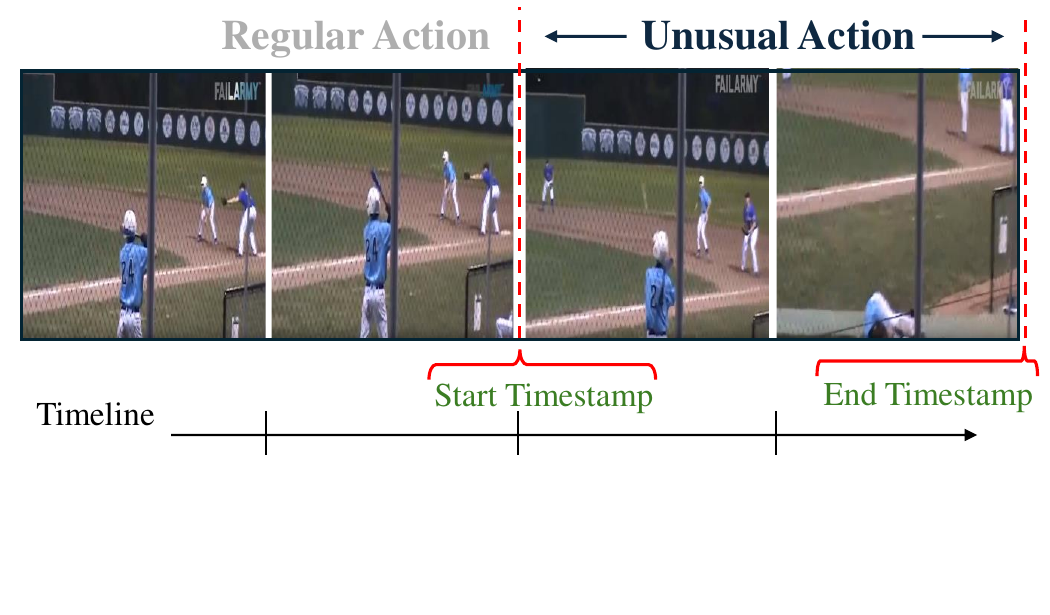}
  \caption{Example of an unusual activity in a baseball game scene. In the 3rd and 4th frames, the ball unexpectedly strikes a batter's head, causing him to fall on the ground. This event is classified as an unusual action.} 
  \vspace{-0.2cm}
  \label{fig: ual_example}
\end{figure}

\noindent In the realm of real-world events and human behavior analysis, unusual activities (as shown in \autoref{fig: ual_example}) are defined as behaviors, patterns, or events that deviate from expected norms or regular occurrences. This concept encompasses two primary aspects: first, outcomes that contradict conventional expectations, exemplified by unintentional \cite{epstein2019oops} and humorous activities \cite{xie2023funqa}; second, rare or infrequent occurrences, such as behaviors related to autism spectrum disorder (ASD) \cite{zahan2023human_gesture_autism}, sudden road accidents \cite{jiao23survey_vad}, natural disasters \cite{acharya2023Disasters}, extreme weather\cite{Finkel2024BringingST, liu2024eric}, unusual public demonstrations like riot \cite{alsedi2017riot}, observation of rare endangered species in urban environments \cite{van2018species}, space events like unexpected meteor dropping \cite{McMullan2023fireball} etc. In this work, we focus on studying unintentional activities, autism-related behaviors, and humorous events. In certain situations, timely detection of these unusual events is crucial as delays can lead to severe consequences~\cite{jin2023emsassist, jin2023emsassistdemo}. To resolve this issue, we need to detect and pinpoint the span of these unusual activities from video clips. Thus, we aim to address this challenge through temporal activity localization via language query, which involves predicting the timestamps of a video segment that semantically matches with the sentence query \cite{Gao2017TALLTA, wang2022video}. 

\label{sec:intro}

\noindent The emergence of Large Language Models (LLMs) and Vision Language Models (VLMs) has facilitated the development of Video-Language-Models (Vid-LLMs), such as VideoChat2 \cite{li2023mvbench}, Video-LLaMA \cite{zhang-etal-2023-videollama}, and Video-ChatGPT \cite{Maaz2023VideoChatGPT}. These models demonstrate remarkable capabilities in video understanding tasks, including activity localization. However, the pretraining datasets of these foundation models \cite{bommasani2021opportunities}, such as ActivityNet \cite{caba2015activitynet}, Charades-STA \cite{Sig2016Charades}, HowTo100M \cite{Miech2019HowTo100MLA}, MSRVTT \cite{Xu2016MSRVTT}, MSVD \cite{chen2011msvd}, DiDeMo \cite{hendricks18didemo}, and WebVid-2M \cite{Bain21webvid2m}, do not adequately represent the unusual activities discussed previously. This insufficient representation could lead to suboptimal performance \cite{parashar2024neglected, liu2024few} for Vid-LLMs to localize such activities. Whether the latest Vid-LLMs and other activity localization approaches are capable of handling unusual activities has not been sufficiently explored yet in the literature. 

\noindent To address this gap, we introduce UAL-Bench, which comprises three benchmark datasets: UAG-OOPS, UAG-SSBD, UAG-FunQA and one instruction-tune dataset: OOPS-UAG-Instruct, designed to enhance model's understanding of unusual activities. UAL-Bench also features a comprehensive evaluation of recent Vid-LLMs and our proposed VLM-LLM approach for unusual activity localization. Additionally, we present a novel metric, $R@1, TD\leq p$, to overcome the limitations of existing metric that have proven unreliable in extreme cases. 

\noindent We explore three approaches to unusual activity localization. Firstly, we evaluate the Vid-LLMs trained on conventional activity datasets mentioned earlier. Secondly, we propose an integration of VLMs and LLMs approach, creating time-aware text representation of videos with VLMs and using LLMs for unusual activity localization in a text-to-text manner. Finally, we fine-tune Vid-LLM using our OOPS-UAG-Instruct dataset and assess its performance on the unusual activity localization task. Our evaluation includes \textbf{three} Vid-LLM models, \textbf{two} VLM-LLM models, and \textbf{one} fine-tuned Vid-LLM model. We identify \textbf{nine} observations that provide a foundation for future researchers to solve unusual activity localization. Notably, our VLM-LLM approach demonstrates improved capabilities to localize extremely short-span unusual events and predicts the onset of any unusual event better than current VideoLLMs. In addition, the VLM-LLM approach exhibits superior reasoning capabilities. Our datasets present similar challenges for the VLM-LLM approach compared to standard benchmark datasets (i.e. Charades-STA\cite{Sig2016Charades}). We also find that long-duration videos, particularly those related to autism diagnosis, pose significant challenges for both Vid-LLM and VLM-LLM approaches. We summarize our major contributions in the following: 
\begin{itemize} 
    \item We propose UAL-Bench, the first comprehensive benchmark for unusual activity localization, which includes three datasets for unusual activity localization: UAG-OOPS, UAG-SSBD, UAG-FunQA.
    \item We present a comprehensive evaluation of three distinct approaches for unusual activity localization: existing Vid-LLMs, a novel integration of Language and Vision models (VLM-LLM), and instruction-tuned Vid-LLMs, providing valuable insights and observations for future research in this area.
    \item We propose a new metric $R@1, TD \leq P$, to address the limitation of existing metric for activity localization task.
    \item We introduce OOPS-UAG-Instruct, an instruction-tuning dataset aimed at enhancing future models' capabilities in understanding and localizing unusual activities.
\end{itemize}
Overall, our work provides a benchmark for unusual activity localization and can potentially be used to improve unusual activity localization ability for existing methodologies. 

\section{Related Works}
\paragraph{Temporal Activity Localization.}
Temporal activity localization seeks to identify complex and diverse activities in videos based on natural language queries. Since its introduction in 2017 \cite{Gao2017TALLTA, lisa2017moment}, this task has preserved ongoing research interest \cite{lan2023survey,zhang2023temporal,gao2021fast} due to its relevance to various video understanding applications, including video summarization, video editing \cite{pramanick2023egovlpv2,tang2022multi}, multimedia information retrieval \cite{zhao2023learning}, healthcare \cite{eysenbach2023role}, and surveillance \cite{de2023socratic} applications. To address the challenges of this task, researchers have proposed several methods focusing on multimodal interactions between videos and sentence queries. The popular methods include proposal-based \cite{he2019read}, proposal-free \cite{zhang2020span}, reinforcement learning based \cite{wu2020tree} and weakly supervised methods \cite{chen2021towards}. Recent advancements feature visual prompting \cite{Bahng2022ExploringVP,Zhang2023TextVisPrompt}, Event Activation Sequence (EAS) based visual-text aligning \cite{Hao2023Fine}. However, these methods rely on annotated training datasets such as TaCoS \cite{regneri2013grounding}, Charades-STA\cite{Sig2016Charades}, ActivityNet-Captions dataset \cite{krishna2017dense}, which primarily focus on cooking, indoor activity and common open-domain activities respectively. Notably, these datasets do not encompass unusual activities\cite{epstein2019oops, xie2023funqa,rajagopalan2013self}.

\paragraph{Large Language Models for temporal video grounding.}
Large language models (LLM) \cite{achiam2023gpt,floridi2020gpt,meta2024introducing} pretrained on extensive datasets demonstrate capabilities in summarization, reasoning, and other tasks \cite{radford2018improving,kojima2022large,min2023recent} without the need for additional fine-tuning. Recent research has explored applications of LLMs in the computer vision domain, particularly in the field of video understanding \cite{tang2023video}. The integration of LLMs allows vision models to analyze and comprehend complex relationships between visual and textual information, leading to the development of Visual Language Models (VLM) \cite{li2023blip2, liu2024visual} for image data and Video Language Models (Vid-LLM) \cite{2023videochat,li2023mvbench,zhang-etal-2023-videollama,Maaz2023VideoChatGPT} for video data. In video understanding, researchers have utilized Vid-LLMs for different tasks including recognition, anticipation, captioning, and temporal localization \cite{li2024multimodal,abdar2023review}. Many studies focus on capturing the temporal aspect of the videos \cite{chen2023video,yang2023vid2seq, zhang-etal-2023-videollama,luo2023valley} through LLM-integrated visual encoders like CLIP\cite{radford2021learning}, BLIP2\cite{li2023blip2}, alongside temporal encoders such as TimesSFormer \cite{bertasius2021space}. However, despite the enhanced reasoning capabilities of LLMs and the availability of the integrated visual encoders, the application of these models to solve temporal video grounding tasks in text-to-text manner \cite{wang2024hawkeye, feng2023llm4vg} along with unusual activities remains largely unexplored .
\section{Problem Formulation}\label{sec: problem formulation}
\noindent\textbf{Task Definition.} An activity localization model takes a video-query pair as input, where the video is a sequence of frames and the query consists of a sequence of words that describes the unusual activity. The primary objective of the method is to identify the temporal boundaries of the visual activity that aligns with the query, specifically the start and end times. Given an untrimmed video $V$, represented as $V = \{f_1,f_2,..., f_T\}$ with $T$ total frames, and a query $Q$, the aim is to identify start timestamp $T_s$ and the end timestamp $T_e$ of the activity within $V$. Thus, the activity localization task can be formally expressed as, $F(V,Q) = (T_s, T_e) $.\\
\textbf{Evaluation protocol.} Activity localization is commonly evaluated using two metrics: $R@n, IoU\geq m$ and $mIoU$, as introduced by Gao et al \cite{Gao2017TALLTA}. The temporal Intersection Over Union ($IoU$) quantifies the overlap between the ground truth start and end times ($g_s, g_e$) and the predicted start and end time ($p_s,p_e$), with values ranging from 0.0 to 1.0. A higher IoU indicates better alignment between segments, while an IoU of 1.0 means an exact match. The calculation of temporal IoU is detailed in \autoref{eq:iou}. 
\begin{equation}\label{eq:iou}
\begin{split}
Intersection & = max(0,min(g_e,p_e)-max(g_s,p_s))\\
Union & = (g_e - g_s) + (p_e-p_s) - Intersection \\
IoU & = \frac{Intersection}{Union}
\end{split}
\end{equation}
The $mIoU$, as illustrated in \autoref{eq: miou}, represents the average of all temporal $IoUs$ derived from the samples $S_i \epsilon \{(g_s,g_e),(p_s,p_e)\}$. 
\begin{equation}\label{eq: miou}
    mIoU = \dfrac{1}{N_s}\sum\limits_{i=1}^{N_s} IoU(S_i)
\end{equation}
The metric $R@n, IoU\geq m$ measures the percentage of samples with at least one prediction whose temporal $IoU$ is equal or greater than $m$ among the top-n predicted timestamps. In our evaluation, conducted in a zero-shot setting with a single prediction per sample, we set $n = 1$. For each sample $S_i \epsilon \{(g_s,g_e),(p_s,p_e)\}$,  if  $IoU\geq m$,  $S_i$ is considered correct. Consequently, $R@1, IoU\geq m$ is computed as outlined in \autoref{recall}:

\begin{equation}\label{recall}
\begin{split}
R@1, IoU\geq & m = \dfrac{1}{N_s}\sum\limits_{i=1}^{N_s} f(n=1,m,S_i) \\
f(n=1,m,S_i) & = 
\begin{cases}
    1,& \text{if } IoU\{(g_s,g_e), (p_s,p_e)\}\geq m\\
    0,              & \text{otherwise}
\end{cases}
\end{split}
\end{equation}

\begin{figure}[t]
  \includegraphics[width=\columnwidth]{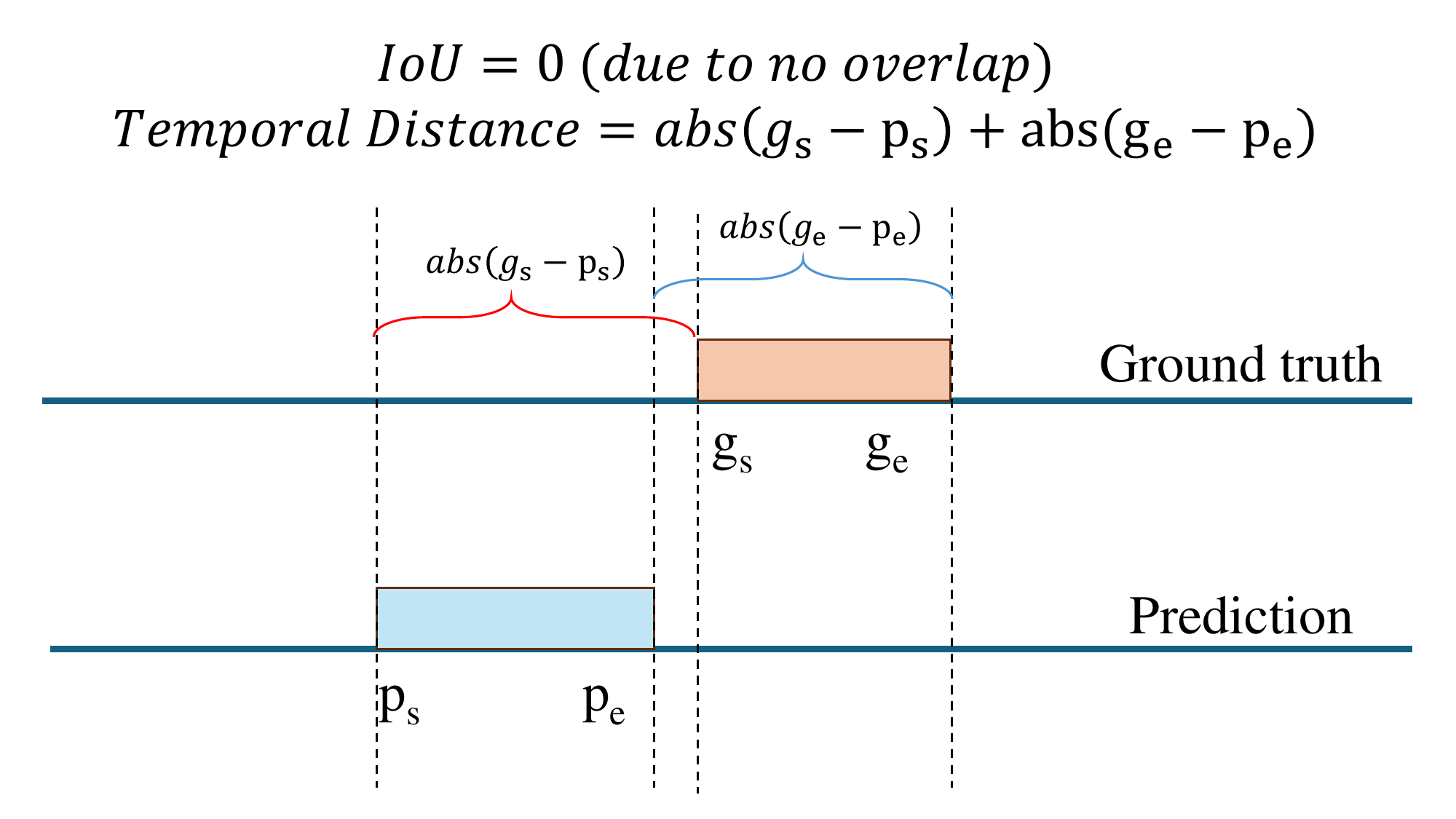}
  \caption{An illustration of our proposed Temporal Distance.}
  \label{fig: td}
  \vspace{-4mm}
\end{figure}

\noindent We also incorporate Temporal Distance (TD) between the ground truth and predicted timestamps as an additional metric to assess the performance of the activity localization task. This consideration arises from the limitation of the $IoU$ metric, which results in a score of 0 when there is no overlap, even if the prediction is very close to the ground truth, as depicted in \autoref{fig: td}. Hence, we compute $R@1, TD \leq p$ where a lower $TD$ indicates better performance and a $TD$ of 0 signifies an exact match. For each sample $S_i$, if the $TD$ is within a threshold of $p$ seconds, it is considered as correct. Therefore, $R@1, TD \leq p$ is calculated as detailed in \autoref{eq:recall-td}: 

\begin{equation}\label{eq:recall-td}
\begin{split}
R@1, TD \leq p & = \dfrac{1}{N_s}\sum\limits_{i=1}^{N_s} g(n=1,p,S_i) \\
g(n=1,p,S_i) & = 
\begin{cases}
    1,& \text{if } TD\{(g_s,g_e), (p_s,p_e)\}\leq p\\
    0,              & \text{otherwise}
\end{cases} \\
\end{split}
\end{equation}

\vspace{-0.6cm}
\section{UAL-Bench: Unusual Activity Localization Bench Mark}\label{sec: ual}

\noindent In this section, we present our proposed UAL-Bench which provides a comprehensive evaluation of three different approaches for unusual activity localization: Vid-LLMs, a novel integration of vision and language models, and fine-tuning pre-trained Video LLMs on task-specific instruction-tune datasets. We describe these approaches in the following paragraphs.

\subsection{Multimodal Vid-LLMs}\label{method:1}
\noindent We employ Vid-LLMs that utilize a video, $V$, and a prompt as inputs to address the localization task. The prompt includes both the query and a description of the unusual activity (as illustrated in \autoref{tab:prompt design}), directing the model to generate the start and end timestamps of the activity. We extract these timestamps from the Vid-LLMs using two methods: pattern matching by regular expressions and manual extraction through analysis of the generated texts. The overall process is depicted in \autoref{eq: videollm}.
\begin{equation} \label{eq: videollm}
\begin{split}
    Output & = Video\_LLM (prompt, V) \\
    T_s, T_e & = Extract(Output)
\end{split}
\end{equation}

\subsection{Integration of Language and Vision Models} \label{method:2}
\noindent We propose a two-step approach for unusual activity localization in a text-to-text manner. In the first step, we generate a time-aware text representation of the video using Visual Language Models (VLMs). In the second step, the time-aware representation is fed into an LLM along with the description of unusual activity to identify the time span in which the activity occurs. This methodology is referred to as VLM-LLM.

\noindent{\bf Time-aware text representation of Videos.} 
To create a time-aware text representation of each video to be used as a context for the LLMs, we first sample each video at a rate of 1 frame per second (fps). This process yields a sequence of sampled frames along with their corresponding timestamps: $Sampled\_Frames(V) = \{(f_1,t_1), (f_2,t_2), \hdots, (f_N,t_N)\}$, where $f_i$ is a frame, $t_i \in T=\{0.0s, 1.0s, 2.0s, \hdots \}$, and $N$ is the total duration of the video in seconds. Subsequently, we iterate through each frame sequentially to generate text descriptions using vision language models (VLM) as demonstrated in \autoref{eq:1}.
\begin{equation}\label{eq:1}
    Desc_{f_i} = VLM (f_i)
\end{equation}
The $VLM (.)$ utilized in our study comprises of two categories: image caption-based models and visual question answering (VQA) models. For our experiments, we selected BLIP-2 \cite{li2023blip2} as the image caption generator and VideoLlaMA \cite{zhang-etal-2023-videollama} as the VQA-based text generator in our experiments. Finally, we concatenate the generated descriptions with the corresponding timestamps sequentially to form a comprehensive text representation of each video, as illustrated in \autoref{eq:2}.
\begin{equation}\label{eq:2}
    Text\_Rep (V) = \{(t_1:Desc_{f_1}), \hdots,(t_N:Desc_{f_N})\}
\end{equation}

\noindent{\bf Video text representation meets Large Language Models.}
In this stage, the LLMs function as video analyst agents tasked with localizing unusual activities and reasoning about the events. The LLM completes the task based on a system instruction and a localization prompt as shown in \autoref{eq:3}. The localization prompt consists of the query and the time-aware text representation of the video. The query is constructed by combining the instruction for the LLM to localize the activity with an annotated description of the activity. All prompt designs are detailed in the Appendix. Finally, we extract the start timestamp ($T_s$) and end timestamp ($T_e$) from the LLM's generated output. We use two different techniques to extract timestamps: pattern matching with regular expressions and manual review of the generated texts. 
\begin{equation}\label{eq:3}
\begin{split}
    Query & = \{Instruction, Activity\_Description(V)\} \\
    Loc\_Prompt & = \{Query, Text\_Rep(V)\} \\
    Output & = LLM\_Agent(Sys\_instruct, Loc\_Prompt) \\
    T_s, T_e & = Extract (Output) 
\end{split}
\end{equation}

\subsection{Instruction-tuning Vid-LLMs}\label{method: 3}
\noindent In this section, we explore the effectiveness of instruction-tuning pre-trained Vid-LLM models for the unusual activity localization task. To facilitate this approach, we have developed the OOPS-UAG-instruct dataset. Detailed information about the instruction-tune dataset is presented in \autoref{sec: exp}. We provide the training videos alongside our instructions in a question-answer format for the pre-trained Vid-LLM models during the fine-tuning process. Subsequently, we evaluate the instruction-tuned models using the test sets of unusual activity datasets, as discussed in detail in \autoref{sec: exp}. 

\section{Experiment Setup and Results}\label{sec: exp}
\subsection{Datasets}
{
\setlength{\tabcolsep}{0.3em}
\begin{table}
  \centering
  \caption{Statistics of the proposed datasets compared to standard temporal localization dataset Charades-STA \cite{Gao2017TALLTA}. Despite being shorter in average duration, OOPS-UAG-Instruct contains more detailed descriptions than Charades-STA.}
  \vspace{-0.2cm}
  {\footnotesize{
  \scalebox{0.95}{
  \begin{tabular}{cccc}
    \toprule
    \multirow{1}{*}{Dataset}  & \# of Videos & \makecell{Avg Duration\\(seconds)} & \makecell{Avg Description length\\(words)}\\
    \midrule
    UAG-OOPS & 1,589 & 8.34
    & 92 \\
    UAG-SSBD & 75 & 90 & 7 \\
    UAG-FunQA & 172 & 7.26 & 5 \\
    OOPS-UAG-Instruct & 3,778 & 9.83& 93.52  \\
    Charades-STA \cite{Gao2017TALLTA} & 3,720 & 30.59 & 33 \\
    \bottomrule
  \end{tabular}
  }}}
  \label{tab:data_table}
\end{table}
}

\noindent In this section, we introduce UAL-Bench Datasets\footnote{\url{https://drive.google.com/drive/folders/1eE_ngd-E6rjdHz0KKttJATzsdxv4Wf_e?usp=sharing}}, which include UAG-OOPS, UAG-SSBD, UAG-FunQA, along with our proposed instruction-tune dataset: OOPS-UAG-instruct. In addition, we employ standard temporal localization dataset, Charades-STA \cite{Sig2016Charades} to evaluate the performance of our proposed VLM-LLM approach across various datasets. The statistics for these datasets are shown in \autoref{tab:data_table}.

\noindent{\bf UAG-OOPS.} The UAG-OOPS dataset is derived from the validation dataset of OOPS! \cite{epstein2019oops}, chosen for its representation of real-world scenarios and human errors \cite{epstein2019oops}. In addition, OOPS! includes the description of primary intention of an activity as well as what went wrong after the failure of that activity. We combined these elements to formulate the descriptions for the UAG OOPS dataset. To determine the start time, we selected the earliest timestamp from the three provided annotations of the activity's start time. As the original dataset does not specify an end time, we designated the video duration as the end timestamp. We excluded all the videos featuring multiple scenes. Additionally, a fourth human annotator's judgment in the original dataset provided a baseline for performance evaluation.\\
{\bf  UAG-SSBD.} The UAG-SSBD dataset is derived from the SSBD \cite{rajagopalan2013self}, comprises of 75 challenging videos featuring children exhibiting self-stimulatory behaviors commonly associated with Autism Spectrum Disorder (ASD). Due to the rarity of such behaviors in mainstream video understanding training datasets, SSBD serves as a unique benchmark. The dataset encompasses three categories of self-stimulatory behaviors, with 25 videos dedicated to each category: arm flapping, head banging, and spinning.  Videos were sourced from YouTube using provided IDs, although only 58 were accessible during experimentation. The UAG-SSBD catalogues a total of 104 behaviors, detailing their start time, end time, and descriptions formatted as: \emph{``A person is [action category] with [intensity] intensity''}. Example: ``A person is spinning with high intensity''.\\
{\bf UAG-FunQA.}
We create UAG-FunQA, a challenging benchmark of counter-intuitive and fun videos from FunQA \cite{xie2023funqa} dataset. FunQA consists of three underexplored video types: Humor, Creative, and Magic. For our study, we selected the test set of HumorQA videos, as these videos are characterized by rapid event changes and the unintentional nature of the depicted activities. To create UAG-FunQA, we selected one of the instructions provided by the dataset (i.e. ``Identify the video's funny moment.'') as description, since the original dataset \cite{xie2023funqa} lacks specific activity descriptions found in other datasets. After that, we converted the start and end frame numbers into timestamps (in seconds) by dividing the frame numbers by 30, given that the videos were sampled at 30 frames per second (FPS).  \\
{\bf OOPS-UAG-instruct dataset.} We developed OOPS-UAG-instruct, an instruction-tune dataset derived from the training set of the OOPS! dataset \cite{epstein2019oops}. The query and answer pairs were constructed following the same format of the VideoChat \cite{2023videochat} instruction data. This dataset comprises 3778 videos along with their corresponding question-answer pairs. Each question provides a description of an activity and asks to predict the start and end time of the activity. The answer section contains the activity segment timestamps. We filtered the single scene training samples from the OOPS! dataset to include only videos containing activities labeled as unintentional and include descriptions.\\
\vspace{-0.5cm}
\subsection{Metrics} 
\noindent We employ three distinct evaluation approaches for localization tasks. Firstly, we use $R@1,IoU \geq m$ to evaluate the unusual activity localization with threshold $m =\{0.3, 0.5, 0.7\}$ following previous studies \cite{Zhang2023TextVisPrompt,feng2023llm4vg,lisa2017moment}. Secondly, to address scenarios where ground truth and predictions do not overlap ($IoU=0$), illustrated in \autoref{fig: td}, we utilize $R@1, TD \leq p$ (\autoref{eq:recall-td}) at thresholds $\{0,1,3,5\}$ and mean temporal distance $mTD$. Lastly, we measure the accuracy of predicting the onset (start time) of unusual activities, following  \cite{epstein2019oops}, within 1 second and 0.25 seconds to emphasize the anticipation of these events.\\
\subsection{Implementation Details}
\noindent This study benchmarks three approaches using the UAG-OOPS, UAG-SSBD, and UAG-FunQA datasets. The baselines include the random method, which predicts start and end timestamps randomly (seed 42) within a range of 0 to 34 seconds (the average duration of the UAG-OOPS, UAG-SSBD, and UAG-FunQA videos combined), and PredictAll \cite{yuan2021closer}, which predicts the entire video as a response. For the UAL-Bench approaches, we implement the Vid-LLM approach using VideoChat2 (videochat2-7b + vicuna-7b-v0) \cite{li2023mvbench}, videoChatGPT (video\_chatgpt-7B + LLaVA-7B-Lightening-v1-1) \cite{Maaz2023VideoChatGPT}, VideoLLaMA (Video-LLaMA-2-7B-Pretrained + llama-2-7b-chat-hf) \cite{zhang-etal-2023-videollama}. The models receive the entire video along with the ``Vid-LLM Prompt'' detailed in \autoref{tab:prompt design}. In the VLM-LLM approach, we utilize  BLIP2 (blip2-opt-2.7b) as the image caption generator and VideoLLaMA (Video-LLaMA-2-7B-Finetuned + llama-2-7b-chat-hf) for VQA-based text generator. For each frame, we pose the question: \emph{``What is happening in the image? Instruction: answer within one line and cover all the details''}. For the LLM, we select LLaMA 3 (Meta-Llama-3-8B-Instruct) \cite{meta2024introducing} due to its extensive context window of 8096 tokens, which is beneficial for processing long videos. In the fine-tuning approach, we use OOPS-UAG-Instruct to fine-tune the vision branch (VL-LLaMA-2-7B-Finetuned.pth) of VideoLlaMA, completing the fine-tuning on a single A100 GPU in approximately 3 hours.
\setlength{\tabcolsep}{0.6em}
\begin{table*}[h]
\centering
\caption{Overall performance comparison of the \colorbox{Apricot!40}{Video-LLM}, \colorbox{BlueGreen!20}{VLM-LLM} and \colorbox{YellowGreen!40}{Fine-tuned VLM} approaches on three unusual activity localization benchmarks: UAG-OOPS, UAG-SSBD and UAG-FunQA. For the $R@1, IoU \geq m$ and $R@1, TD \leq p$ metrics, higher scores indicate better performance, while for the $mTD$ metric, the lower scores are better.}
  \footnotesize
  \begin{tabular}{c|ccccc||ccccc}
    \toprule
    & \multirow{2}{*}{Methods} & \multicolumn{3}{c}{$R@1, IoU\geq m$} & \multirow{2}{*}{$mIoU (0-1)$} & \multicolumn{4}{c}{$R@1,TD\leq p (sec)$} & \multirow{2}{*}{$mTD(sec)$} \\
    & & m=0.3 & m=0.5 & m=0.7 & & p=0 & p=1 & p=3 & p=5 & \\
    \midrule 
    \multirow{7}{*}{\rotatebox[]{90}{UAG-OOPS}} & \cellcolor{Apricot!40} VideoChat2 \cite{li2023mvbench} & \cellcolor{Apricot!40} 16.49 & \cellcolor{Apricot!40} 5.98 & \cellcolor{Apricot!40}1.95 & \cellcolor{Apricot!40}0.12 &\cellcolor{Apricot!40} 0.00 & \cellcolor{Apricot!40}1.01 & \cellcolor{Apricot!40}8.37 & \cellcolor{Apricot!40}23.10 & \cellcolor{Apricot!40}9.31\\ 
    & \cellcolor{Apricot!40}Video-ChatGPT \cite{Maaz2023VideoChatGPT} & \cellcolor{Apricot!40}25.49 & \cellcolor{Apricot!40}10.70 & \cellcolor{Apricot!40}3.15 & \cellcolor{Apricot!40}0.18 & \cellcolor{Apricot!40}0.00 & \cellcolor{Apricot!40}1.32 & \cellcolor{Apricot!40}8.37 & \cellcolor{Apricot!40}25.61 & \cellcolor{Apricot!40}11.50 \\ 
     & \cellcolor{Apricot!40}VideoLLaMA \cite{zhang-etal-2023-videollama} &
     \cellcolor{Apricot!40}\bf40.72 & \cellcolor{Apricot!40}\bf20.77 & \cellcolor{Apricot!40}\bf6.23 & \cellcolor{Apricot!40}\bf0.27 & \cellcolor{Apricot!40}\bf0.06 & \cellcolor{Apricot!40}\bf2.01 & \cellcolor{Apricot!40}14.85 & \cellcolor{Apricot!40}33.79 & \cellcolor{Apricot!40}11.22\\ 
    & \cellcolor{BlueGreen!20} BLIP2-LlaMA3 \cite{li2023blip2,meta2024introducing}(ours) 
    & \cellcolor{BlueGreen!20}19.07 & \cellcolor{BlueGreen!20}7.17 &\cellcolor{BlueGreen!20} 2.45 & \cellcolor{BlueGreen!20}0.15 &\cellcolor{BlueGreen!20} 0.00 & \cellcolor{BlueGreen!20}1.38 & \cellcolor{BlueGreen!20}\bf 27.00 & \cellcolor{BlueGreen!20}53.74 & \cellcolor{BlueGreen!20}5.85  \\
    &\cellcolor{BlueGreen!20} VideoLlaMA-LlaMA3 \cite{zhang-etal-2023-videollama,meta2024introducing} (ours) 
    & \cellcolor{BlueGreen!20}19.38 &\cellcolor{BlueGreen!20}7.93 &\cellcolor{BlueGreen!20} 2.08 & \cellcolor{BlueGreen!20}0.15 &\cellcolor{BlueGreen!20} 0.00 & \cellcolor{BlueGreen!20}1.89 & \cellcolor{BlueGreen!20}26.12 & \cellcolor{BlueGreen!20}\bf55.13 &\cellcolor{BlueGreen!20} \bf5.72\\ 
    &\cellcolor{YellowGreen!40} VideoLlaMA 7B (Fine-tuned) 
    &\cellcolor{YellowGreen!40} 2.96 & \cellcolor{YellowGreen!40}0.50 &\cellcolor{YellowGreen!40} 0.19 &\cellcolor{YellowGreen!40} 0.04 &\cellcolor{YellowGreen!40} 0.00 & \cellcolor{YellowGreen!40}1.26 &\cellcolor{YellowGreen!40} 10.07 & \cellcolor{YellowGreen!40}22.84 & \cellcolor{YellowGreen!40}14.09\\
    \cmidrule(l{3em}r{2em}){2-11} 
    & Random & 12.21  & 4.47 & 1.45 & 0.10 & 0.00 & 0.31 & 2.77 & 5.29 & 24.10\\
    \midrule
     \multirow{7}{*}{\rotatebox[]{90}{UAG-SSBD }} & \cellcolor{Apricot!40} VideoChat2 
     &  \cellcolor{Apricot!40}2.88  & \cellcolor{Apricot!40}0.96 & \cellcolor{Apricot!40}0.00 & \cellcolor{Apricot!40}0.02  &\cellcolor{Apricot!40} 0.00 & \cellcolor{Apricot!40}0.00  &\cellcolor{Apricot!40} 1.92 &\cellcolor{Apricot!40} 2.88 & \cellcolor{Apricot!40}139.63  \\ 
    &\cellcolor{Apricot!40} Video-ChatGPT & \cellcolor{Apricot!40}4.81  & \cellcolor{Apricot!40}2.88 &\cellcolor{Apricot!40} 0.00 & \cellcolor{Apricot!40}0.03 & \cellcolor{Apricot!40}0.00  &\cellcolor{Apricot!40}0.00 & \cellcolor{Apricot!40}0.96 & \cellcolor{Apricot!40}2.88 & \cellcolor{Apricot!40}93.99\\ 
    &\cellcolor{Apricot!40}VideoLlaMA & \cellcolor{Apricot!40}\bf15.38 & \cellcolor{Apricot!40}\bf 8.65 & \cellcolor{Apricot!40}1.92 & \cellcolor{Apricot!40}\bf 0.11 & \cellcolor{Apricot!40}0.00 & \cellcolor{Apricot!40}0.00 & \cellcolor{Apricot!40}\bf3.85 & \cellcolor{Apricot!40}\bf6.73 & \cellcolor{Apricot!40}96.55\\ 
    &\cellcolor{BlueGreen!20} BLIP2-LlaMA3 (ours) & \cellcolor{BlueGreen!20}1.92 & \cellcolor{BlueGreen!20}1.92 &\cellcolor{BlueGreen!20} 1.92 & \cellcolor{BlueGreen!20}0.03 & \cellcolor{BlueGreen!20}0.00  &\cellcolor{BlueGreen!20}0.00 & \cellcolor{BlueGreen!20}0.96 &\cellcolor{BlueGreen!20}1.92&\cellcolor{BlueGreen!20}\bf68.05 \\ 
    &\cellcolor{BlueGreen!20}  VideoLlaMA-LlaMA3 (ours) & \cellcolor{BlueGreen!20}2.88 & \cellcolor{BlueGreen!20}0.96 &\cellcolor{BlueGreen!20} 0.00 & \cellcolor{BlueGreen!20}0.03 &\cellcolor{BlueGreen!20} 0.00 & \cellcolor{BlueGreen!20}0.00 &\cellcolor{BlueGreen!20} 0.96 & \cellcolor{BlueGreen!20}4.81 & \cellcolor{BlueGreen!20}70.38\\ 
    &\cellcolor{YellowGreen!40}  VideoLlaMA 7B (Fine-tuned) &\cellcolor{YellowGreen!40} 0.00 &\cellcolor{YellowGreen!40} 0.00 & \cellcolor{YellowGreen!40}0.00 &\cellcolor{YellowGreen!40} 0.01 & \cellcolor{YellowGreen!40}0.00 &\cellcolor{YellowGreen!40} 0.00 &\cellcolor{YellowGreen!40} 0.00 & \cellcolor{YellowGreen!40}1.92 &\cellcolor{YellowGreen!40} 105.27\\                
    \cmidrule(l{3em}r{2em}){2-11} & Random & 10.58 & 5.77 & \bf3.85 & 0.10 & 0.00 & 0.00 & 1.92 & 1.92 & 87.73\\ 
    \midrule
      \multirow{7}{*}{\rotatebox[]{90}{UAG-FunQA}} & \cellcolor{Apricot!40}VideoChat2 & \cellcolor{Apricot!40}12.79 & \cellcolor{Apricot!40}4.65 & \cellcolor{Apricot!40}3.49 & \cellcolor{Apricot!40}0.08 & \cellcolor{Apricot!40}0.00 &\cellcolor{Apricot!40} 2.33 &\cellcolor{Apricot!40} 23.84 &\cellcolor{Apricot!40} 44.77 &\cellcolor{Apricot!40} 7.48\\
    &\cellcolor{Apricot!40}  Video-ChatGPT & \cellcolor{Apricot!40}1.16 & \cellcolor{Apricot!40}0.58 & \cellcolor{Apricot!40}0.00 & \cellcolor{Apricot!40}0.01 & \cellcolor{Apricot!40}0.00 & \cellcolor{Apricot!40}0.00 & \cellcolor{Apricot!40}22.67 & \cellcolor{Apricot!40}44.77 & \cellcolor{Apricot!40}53.42\\ 
    &\cellcolor{Apricot!40}  VideoLlaMA & \cellcolor{Apricot!40}2.91 & \cellcolor{Apricot!40}0.58 &\cellcolor{Apricot!40} 0.00 &\cellcolor{Apricot!40} 0.02 &\cellcolor{Apricot!40} 0.00 & \cellcolor{Apricot!40}0.00 & \cellcolor{Apricot!40}4.07 & \cellcolor{Apricot!40}8.72 & \cellcolor{Apricot!40}31.64\\ 
    &\cellcolor{BlueGreen!20}  BLIP2-LlaMA3 (ours) &\cellcolor{BlueGreen!20} \bf 18.60 &	\cellcolor{BlueGreen!20}\bf9.30 & \cellcolor{BlueGreen!20}\bf5.23 &\cellcolor{BlueGreen!20}\bf 0.12& \cellcolor{BlueGreen!20}0.00&\cellcolor{BlueGreen!20}\bf9.30&\cellcolor{BlueGreen!20} 39.53 & \cellcolor{BlueGreen!20}60.47 &  \cellcolor{BlueGreen!20}5.43\\ 
    &\cellcolor{BlueGreen!20}\cellcolor{BlueGreen!20}  VideoLlaMA-LlaMA3 (ours) &\cellcolor{BlueGreen!20} 12.21 & \cellcolor{BlueGreen!20}4.65	& \cellcolor{BlueGreen!20}2.33&\cellcolor{BlueGreen!20}	0.09 &\cellcolor{BlueGreen!20} 0.00 & \cellcolor{BlueGreen!20}5.23&\cellcolor{BlueGreen!20}\bf44.19 &\cellcolor{BlueGreen!20}\bf65.70 & \cellcolor{BlueGreen!20}\bf4.93 \\ 
    &\cellcolor{YellowGreen!40}  VideoLlaMA 7B (Fine-tuned) & \cellcolor{YellowGreen!40}6.40 & \cellcolor{YellowGreen!40}2.33 & \cellcolor{YellowGreen!40}0.0 &\cellcolor{YellowGreen!40}0.01 & \cellcolor{YellowGreen!40}0.00 & \cellcolor{YellowGreen!40}5.81 & \cellcolor{YellowGreen!40}29.65 & \cellcolor{YellowGreen!40}47.67 &\cellcolor{YellowGreen!40} 8.19\\ 
    \cmidrule(l{3em}r{2em}){2-11} 
    & Random & 5.81 & 1.74 & 0.58 & 0.05 & 0.00 & 0.00 & 1.74 & 4.65 & 27.27\\ 
    \bottomrule
  \end{tabular}
  \label{tab:main table}
\end{table*}

\subsection{Results}  
\noindent We present our results organized by three datasets, reflecting their distinct characteristics, including average duration and event types. Each dataset features a comparison of our three UAL-Bench approaches, with overall performance on unusual activity localization tasks summarized in \autoref{tab:main table}.\\
{\bf UAG-OOPS.} Overall scores for this dataset are low across both metrics, particularly at thresholds (m = 0.7, p = 0.1), indicating the challenges in identifying unusual activities. The Vid-LLM approach outperformed others, with VideoLlaMA \cite{zhang-etal-2023-videollama} achieving the highest performance. The two VLM-LLM models exhibited similar results, achieving better scores at higher thresholds of $p = 3, 5$ and $mTD$. Notably, the fine-tuning approach did not yield better results than random method at certain thresholds of $IoU$.\\
{\bf UAG-SSBD.} This dataset poses the greatest challenges, with all approaches yielding lower overall scores compared to others. VideoLlaMA\cite{zhang-etal-2023-videollama} achieves superior performance at thresholds $m=0.3, 0.5$ and $p = 3, 5$. However, at thresholds ($m=0.7, p=0,1$) that are closer to the original segment, all the approaches performed equally or worse than the random baseline, with the VLM-LLM and fine-tuning approaches particularly underperforming.\\
{\bf UAG-FunQA.} The VLM-LLM approach exhibited enhanced capabilities in managing extremely short and humorous videos. BLIP2-LlaMA3 \cite{li2023blip2,meta2024introducing} achieved the best results across all thresholds, particularly at the high-quality threshold of $p = 1$ for $TD$. Among Vid-LLMs, VideoChat2\cite{li2023mvbench} was particularly effective in localizing funny moments in these short videos. In addition, the fine-tuning approach has also shown better performance compared to other datasets.
\begin{figure}[t]
  \includegraphics[scale=0.46]{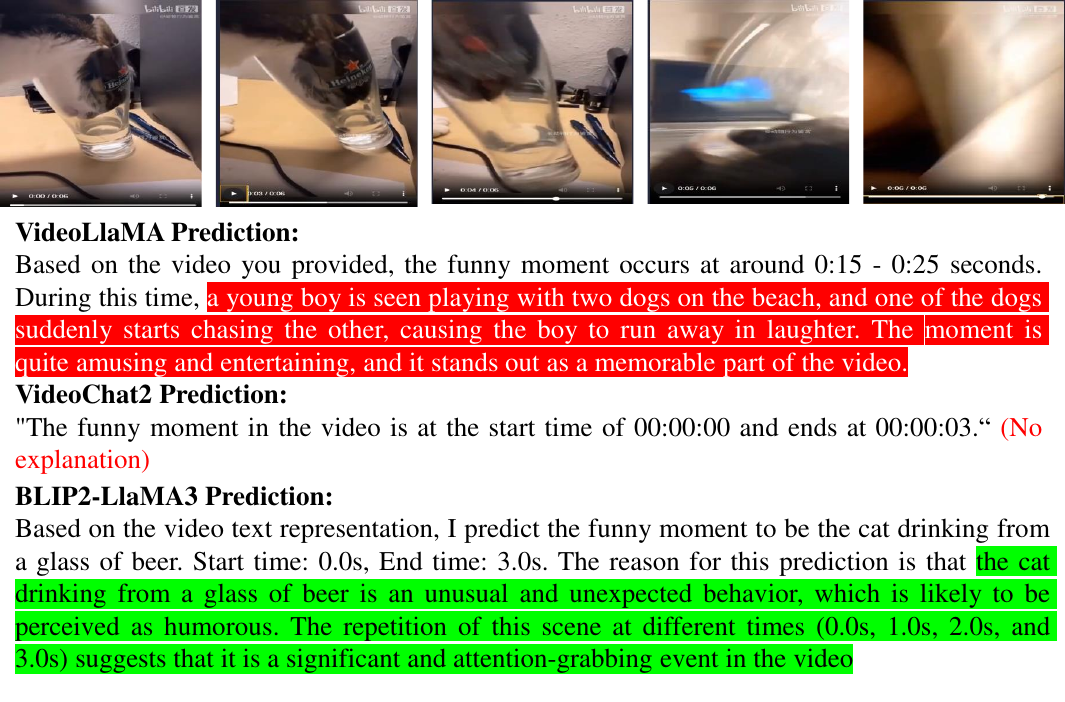}
  \caption{Comparison of explanations among the best-performing models from our experiments. VLM-LLM approach demonstrates a superior understanding of the scene compared to other models. Explanations highlighted in Red highlight indicates incorrect, while those in green signify the correct interpretation.}
  \label{fig:llm_prompt_explanation}
\end{figure}

\section{Discussions}\label{sec: discuss}
\setlength{\tabcolsep}{0.6em}
\begin{table}[h]
  \centering
  \caption{Performance comparison of \colorbox{Apricot!40}{Vid-LLMs}, \colorbox{BlueGreen!20}{VLM-LLM} and \colorbox{YellowGreen!40}{Fine-tuned VLM} for localizing the onset of unusual activity. VLM-LLM approach shows primary abilities to localize short-span unusual events.}
  \vspace{-0.2cm}
  \scalebox{1.0}{\footnotesize{
  \begin{tabular}{c|c|p{1cm}|p{1cm}}
    \toprule
    & \centering \multirow{3}{*}{Methods}& Accuracy within 1 sec & Accuracy within 0.25 sec \\ 
    \midrule 
     \multirow{9}{*}{\rotatebox[]{90}{UAG-OOPS}} &\cellcolor{Apricot!40} VideoChat2 & \cellcolor{Apricot!40}10.01 &\cellcolor{Apricot!40} 3.90\\ 
     & \cellcolor{Apricot!40}Video-ChatGPT &\cellcolor{Apricot!40} 9.38&\cellcolor{Apricot!40}3.90\\ 
     &\cellcolor{Apricot!40} VideoLlaMA & \cellcolor{Apricot!40}16.11&\cellcolor{Apricot!40}4.85\\ 
     & \cellcolor{BlueGreen!20}BLIP2-LlaMA3 &\cellcolor{BlueGreen!20} \bf39.33& \cellcolor{BlueGreen!20}\bf10.32\\ 
     & \cellcolor{BlueGreen!20}VideoLlaMA-LlaMA3 &\cellcolor{BlueGreen!20}38.07& \cellcolor{BlueGreen!20}9.19\\ 
     &\cellcolor{YellowGreen!40} VideoLlaMA 7B (Fine-tuned)& \cellcolor{YellowGreen!40}23.22& \cellcolor{YellowGreen!40}5.60 \\
    & Random & 15.29 & 5.48\\ 
    & Video-Speed (self-supervised)~\cite{epstein2019oops} &65.3&36.6 \\
    & Human Prediction & 72.50 & 25.17 \\ 
    \midrule \multirow{6}{*}{\rotatebox[]{90}{UAG-SSBD }} & \cellcolor{Apricot!40}
\cellcolor{Apricot!40}VideoChat2 &\cellcolor{Apricot!40}
\cellcolor{Apricot!40} 6.73 &\cellcolor{Apricot!40}
\cellcolor{Apricot!40} 0.00\\ 
    &\cellcolor{Apricot!40}  Video-ChatGPT &\cellcolor{Apricot!40}
\cellcolor{Apricot!40} 7.69 &\cellcolor{Apricot!40}
\cellcolor{Apricot!40} 0.00\\ 
    &\cellcolor{Apricot!40}  VideoLlaMA & \cellcolor{Apricot!40}
\cellcolor{Apricot!40}\bf15.38 & \cellcolor{Apricot!40}
\cellcolor{Apricot!40}\bf2.88\\ 
    & \cellcolor{BlueGreen!20}  BLIP2-LlaMA3 & \cellcolor{BlueGreen!20}4.81 & \cellcolor{BlueGreen!20}0.00\\ 
     & \cellcolor{BlueGreen!20}VideoLlaMA-LlaMA3 & \cellcolor{BlueGreen!20}8.65 &	\cellcolor{BlueGreen!20}\bf2.88\\ 
    &\cellcolor{YellowGreen!40}  VideoLlaMA 7B (Fine-tuned) &\cellcolor{YellowGreen!40} \cellcolor{YellowGreen!40}11.54 &\cellcolor{YellowGreen!40} 1.92 \\
    & Random & 2.88	& 1.92 \\ 
     \midrule \multirow{6}{*}{\rotatebox[]{90}{UAG-FunQA }} & \cellcolor{Apricot!40}VideoChat2 & \cellcolor{Apricot!40}\bf50.00 & \cellcolor{Apricot!40}\bf31.98\\ 
    &\cellcolor{Apricot!40} Video-ChatGPT & \cellcolor{Apricot!40}1.74 & \cellcolor{Apricot!40}1.74\\ 
     &\cellcolor{Apricot!40} VideoLlaMA & \cellcolor{Apricot!40}3.49 & \cellcolor{Apricot!40}2.33\\ 
     &\cellcolor{BlueGreen!20} BLIP2-LlaMA3 & \cellcolor{BlueGreen!20}38.37 &\cellcolor{BlueGreen!20}	13.95\\ 
    &\cellcolor{BlueGreen!20} VideoLlaMA-LlaMA3 & \cellcolor{BlueGreen!20}41.28 & \cellcolor{BlueGreen!20}10.47\\ 
     &\cellcolor{YellowGreen!40} VideoLlaMA 7B (Fine-tuned) & \cellcolor{YellowGreen!40}36.05 &\cellcolor{YellowGreen!40}s 8.14 \\
    & Random & 8.72 & 3.49 \\
    \bottomrule
  \end{tabular}
  }}
  \label{tab:anticipation}
\end{table}
\noindent In this section, we summarize our findings and present \textbf{nine} observations derived from the results. These observations are as follows:

\noindent {\bf VLM-LLM approach excels in localizing short-span unusual events}. The VLM-LLM approach significantly outperforms existing Vid-LLMs in short-span video datasets, specifically UAG-OOPS and UAG-FunQA. Our analysis of onset (start time) localization for unusual events detailed in \autoref{tab:anticipation}, reveals that although human predictions exceed those of VLM-LLM in the UAG-OOPS dataset, the VLM-LLM consistently surpasses current Vid-LLMs across both short-span datasets. Furthermore, We compare our zero-shot performances with video-Speed \cite{epstein2019oops}, a self-supervised model trained on OOPS training split. The significant accuracy improvement over Vid-LLMs in zero-shot settings for OOPS dataset, highlights the challenges of applying foundation video models to unusual activity localization task.

\noindent {\bf Explainability: VLM-LLM models provide superior explanations of the predictions.}  We assessed the accuracy of explanations generated by these models for their predictions by comparing the outputs of the best-performing models across all datasets. As shown in \autoref{fig:llm_prompt_explanation}, the VLM-LLM not only offers accurate explanations but also effectively identifies repeated scenes in the sampled frames.

\noindent{\bf Our VLM-LLM approach performs promisingly in standard temporal activity localization  benchmark.}
 We conducted a zero-shot evaluation of the VLM-LLM approach across our datasets and the standard Charades-STA dataset \cite{Sig2016Charades}. As indicated in \autoref{tab:benchmark_comp}, our VLM-LLM outperforms most popular video LLMs, with the exception of Video-ChatGPT-7B.

\noindent {\bf Our datasets challenge the VLM-LLM approach comparatively similar to Charades-STA.}
\autoref{tab:benchmark_comp} demonstrates that VLM-LLM approach performs almost similarly in UAG-OOPS, UAG-FunQA and Charades-STA \cite{Sig2016Charades} benchmarks, though it shows weaker performance in UAG-SSBD. This indicates that our benchmarks are as challenging as standard localization benchmarks.

\noindent{\bf The $R@1,IoU \geq m$ metric becomes unreliable for short-span videos}. In datasets like UAG-OOPS and UAG-FunQA, where entire short videos are often annotated as unusual, setting a low threshold  (e.g., $m = 0.3$) allows a naive prediction of the entire video (PredictAll baseline) as unusual to surpass the performance of top models (\autoref{tab:iou_issue}). However, UAG-SSBD, with its longer videos, reports lower scores for the PredictAll baseline. This highlights the need for metrics tailored to short video events lasting only seconds to ensure accurate model evaluation. 

\noindent{\bf Trade-offs between model complexity (computing cost) and VLM-LLM performance.}
The novel VLM-LLM approach introduces added complexity to the model architecture, necessitating an analysis of computing cost (inference time) per video compared to Vid-LLMs. As shown in \autoref{tab:infer_performance_table}, generating video-text descriptions via VLMs is more time-intensive, as it requires sampling the video into several frames and processing each frame individually. Among the tested VLMs, VideoLlaMA offers faster inference. However, in terms of performance, we observe a two-fold increase in accuracy for unusual activity detection in the UAG-OOPS benchmark.

\noindent {\bf Long duration diagnosis videos (UAG-SSBD) require greater attention}. The UAG-SSBD dataset, one of our benchmarks, differs significantly from the others. It consists of videos capturing various self-stimulatory behaviors in autistic children, which are longer in duration and can be used for diagnostic purposes by localizing behaviors and identifying their onset. Our findings indicate that both Vid-LLMs and VLM-LLM approaches marginally outperform random baseline methods. One reason for this is that the training datasets used for Vid-LLMs, consisting of video-text or image-text pairs, do not prioritize diagnosis-oriented content like that in UAG-SSBD. In the case of VLM-LLM, the long video durations result in lengthy video-text representations, which challenge the LLMs' ability to fully interpret the context, even when using advanced models like LlaMA3 \cite{meta2024introducing} with an 8096-token capacity. Additionally, video text generator models are also not trained to diagnose behavior-related content. However, the latest LLMs are very good at understanding cues from the text. Compared to Vid-LLMs, VLM-LLM methods are very good at explaining their answers after giving the prediction even if we do not explicitly ask them to explain. With these facts aligned, more robust approaches are needed to interpret unpredictable, long-duration diagnostic videos such as self-stimulatory behaviors. 

\noindent{\bf Instruction-tuning issue}. We observe that instruction-tuning consistently underperformed, often producing results worse than random predictions. 
We attribute this performance decline to a key architectural limitation: the lack of a time-awareness module in the video encoder. The model does not encode timestamp information for video frames during pretraining or fine-tuning. Specifically, VideoLlaMA samples eight frames uniformly from the input video, irrespective of its length. These frames are then encoded into tensors and passed to VideoQformer for inter-frame context learning. As a result, when time-specific instructions are provided during fine-tuning or inference, the model lacks access to critical timestamp information, leading to decreased performance. We propose that incorporating a time-awareness module in the video encoder could address this issue and leave this for future work.

\noindent{\bf Explicit content detection and model refusals}. The UAG-OOPS dataset contains explicit annotations (e.g., ``hit the child'', ``dead'', ``shoot a gun'') which triggered ethical guidelines, causing VideoLlaMA to refuse inference. Similarly, LLaMA3 \cite{meta2024introducing} refused to make predictions due to explicit language in video-text representations, particularly those generated by BLIP2 \cite{li2023blip2}. This suggests that careful word choice in annotating unusual activities is essential to avoid such refusals.

\noindent{\bf Limitations and Future work.}
We acknowledge that the VLM-LLM approach primarily works for short videos. Additionally, the original OOPS! \cite{epstein2019oops} dataset lacks end-time annotations of each activity. Since the average duration of UAG-OOPS videos is short (8.34 seconds), we assumed the finishing time of the videos as the end-time annotation in this study. Another limitation is that our study does not consider audio or acoustic data, which may impact performance. Furthermore, the domain-specific nature of our dataset could affect generalizability of our methods. To address these limitations, we propose several future directions: expanding the domains, such as broadening the autism dataset; integrating time-aware encodings into Vid-LLM models to leverage our instruction-tune dataset; and generating end-time annotations for the UAG-OOPS dataset. These efforts aim to enhance the overall quality of the datasets and advance state-of-the-art video understanding models. 
\setlength{\tabcolsep}{0.3em}
\begin{table}[h]
  \centering
  \caption{Comparison of VLM-LLM approach performance between our benchmarks and widely-used Charades-STA \cite{Sig2016Charades}. Our datasets present challenges to the VLM-LLM approach comparable to those in Charades-STA. Additionally, a comparison between the VLM-LLM and Video-LLM models on Charades-STA reveals that the VLM-LLM approach outperforms  Vid-LLMs on the Charades-STA benchmark.}
  \vspace{-0.2cm}
  {\footnotesize{
  \scalebox{1.0}{
  \begin{tabular}{p{1.5cm}p{3.5cm}p{0.6cm}p{0.6cm}p{0.6cm}}
    \toprule
    \multirow{2}{*}{Dataset} & \multirow{2}{*}{Models} & \multicolumn{3}{c}{$R@1, IoU\geq m$} \\
    & & m=0.3 & m=0.5 & m=0.7 \\
    \midrule
    \multirow{3}{*}{UAG-OOPS} & BLIP2-LlaMA3 \cite{li2023blip2,meta2024introducing} & 19.07 & 7.17 & \bf2.45 \\
    & VideoLlaMA-LlaMA3 \cite{zhang-etal-2023-videollama,meta2024introducing} & \bf19.38 &	\bf7.93 & 2.08\\ 
    \midrule
    \multirow{2}{*}{UAG-SSBD} & BLIP2-LlaMA3 & 1.92 & 1.92 & \bf1.92 \\ 
    & VideoLlaMA-LlaMA3 & \bf2.88 & \bf0.96 & 0.00\\                          
    \midrule
    \multirow{2}{*}{UAG-FunQA} & BLIP2-LlaMA3 & \bf18.60 & \bf9.30 & \bf5.23 \\ 
    & VideoLlaMA-LlaMA3 & 12.21 & 4.65 & 2.33\\ 
    \midrule
    \multirow{5}{*}{Charades-STA } & BLIP2-LlaMA3 & 15.00 & 6.45 & \bf1.72\\ 
    & VideoLlaMA-LlaMA3 & 11.24 & 4.19 & 1.10\\
    &VideoChat-7B \cite{2023videochat}\cite{zheng2024training} & 3.0 & 3.3 & 1.3 \\
    &VideoLlaMA-7B \cite{zhang-etal-2023-videollama}\cite{zheng2024training} &10.4&3.8&0.9\\
    &Video-ChatGPT-7B \cite{Maaz2023VideoChatGPT}\cite{zheng2024training} & \bf20.0 &\bf7.7 &1.7\\
    \bottomrule
  \end{tabular}
  }}}
  \label{tab:benchmark_comp}
\end{table}
\setlength{\tabcolsep}{0.3em}
\begin{table}[h]
  \centering
  \caption{Unreliablity of $R@1,IoU \geq m$ at lower thresholds(e.g. 0.3) for short videos where annotations of unusual activity often span the entire video duration.}
  \vspace{-0.2cm}
  \footnotesize
  \scalebox{1.0}{
  \begin{tabular}{p{1.5cm}p{3.0cm}p{0.7cm}p{0.7cm}p{0.7cm}}
    \toprule
    \multirow{2}{*}{Dataset} & \multirow{2}{*}{Methods} & \multicolumn{3}{c}{$R@1, IoU\geq m$}  \\
    &  & m=0.3 & m=0.5 & m=0.7\\
    \midrule
    \multirow{2}{*}{UAG-OOPS}& VideoLLaMA & 40.72 & 20.77 & 6.23 \\ 
    & PredictAll & \bf85.71 & \bf64.88 & \bf31.40 \\
    \midrule
    \multirow{2}{*}{UAG-SSBD}&VideoLlaMA & 15.38 & 8.65 & 1.92 \\
    & PredictAll & \bf25.00 & \bf14.42 & \bf6.73\\ 
    \midrule
    \multirow{2}{*}{UAG-FunQA}& BLIP2-LlaMA3 (ours) & 18.60 & 9.30 & 5.23 \\ 
    & PredictAll & \bf81.40 & \bf58.72 & \bf41.86\\
    \bottomrule
  \end{tabular}
  }
  \label{tab:iou_issue}
\end{table}
\setlength{\tabcolsep}{0.3em}
\begin{table}[h]
  \centering
  \caption{Inference time vs performance trade-off between our VLM-LLM approach and Vid-LLM. A 2X improvement in accuracy is observed with the increased complexity of our approach in unusual activity detection task on UAG-OOPS benchmark.}
  \vspace{-0.2cm}
  {\small{
  \scalebox{0.83}{
  \begin{tabular}{ccccp{1.6cm}}
    \toprule
    \multirow{1}{*}{Method}& Model  & GPU & Time per Video & Accuracy within 1 sec\\
    \midrule
    Video-LLM & VideoLlaMA& 24GB &7s & 16.11\\
    \midrule
    \multirow{2}{*}{\makecell{VLM-LLM}} & BLIP2+LlaMA3& 24GB &25s + 10 s&39.33\\
    
       &VideoLlaMA+LlaMA3 &24GB&24s +10s &38.07\\
    \bottomrule
  \end{tabular}
  }}}
  \label{tab:infer_performance_table}
\end{table}

\section{Conclusions}
\noindent In this paper, we propose UAL-Bench, a comprehensive benchmark designed to evaluate three distinct methods across our proposed datasets for unusual activity localization: UAG-OOPS, UAG-SSBD, and UAG-FunQA. Additionally, We propose a new metric, $R@1, TD\leq p$, to address the limitation of existing metrics, which have proven unreliable in certain scenarios. Our results indicate that the proposed VLM-LLM approach effectively localizes extremely short-span unusual activities while providing more robust explanations. However, we also found that long-duration autism diagnosis videos pose significant challenges for current methods, highlighting the need for further advancements in this area.

\clearpage

{\small
\bibliographystyle{ieee_fullname}
\bibliography{egbib}

\begin{thebibliography}{10}\itemsep=-1pt

\bibitem{abdar2023review}
Moloud Abdar, Meenakshi Kollati, Swaraja Kuraparthi, Farhad Pourpanah, Daniel McDuff, Mohammad Ghavamzadeh, Shuicheng Yan, Abduallah Mohamed, Abbas Khosravi, Erik Cambria, et~al.
\newblock A review of deep learning for video captioning.
\newblock {\em arXiv preprint arXiv:2304.11431}, 2023.

\bibitem{acharya2023Disasters}
{Viral V.} Acharya, {Timothy C} Johnson, {Suresh M.} Sundaresan, and Steven Zheng.
\newblock Disasters with unobservable duration and frequency: Intensified responses and diminished preparedness.
\newblock WorkingPaper 31067, National Bureau of Economic Research, Mar. 2023.

\bibitem{achiam2023gpt}
Josh Achiam, Steven Adler, Sandhini Agarwal, Lama Ahmad, Ilge Akkaya, Florencia~Leoni Aleman, Diogo Almeida, Janko Altenschmidt, Sam Altman, Shyamal Anadkat, et~al.
\newblock Gpt-4 technical report.
\newblock {\em arXiv preprint arXiv:2303.08774}, 2023.

\bibitem{alsedi2017riot}
Nasser Alsaedi, Pete Burnap, and Omer Rana.
\newblock Can we predict a riot? disruptive event detection using twitter.
\newblock {\em ACM Trans. Internet Technol.}, 17(2), mar 2017.

\bibitem{Bahng2022ExploringVP}
Hyojin Bahng, Ali Jahanian, Swami Sankaranarayanan, and Phillip Isola.
\newblock Exploring visual prompts for adapting large-scale models.
\newblock 2022.

\bibitem{Bain21webvid2m}
Max Bain, Arsha Nagrani, G{\"u}l Varol, and Andrew Zisserman.
\newblock Frozen in time: A joint video and image encoder for end-to-end retrieval.
\newblock In {\em IEEE International Conference on Computer Vision}, 2021.

\bibitem{bertasius2021space}
Gedas Bertasius, Heng Wang, and Lorenzo Torresani.
\newblock Is space-time attention all you need for video understanding?
\newblock In {\em ICML}, volume~2, page~4, 2021.

\bibitem{bommasani2021opportunities}
Rishi Bommasani, Drew~A Hudson, Ehsan Adeli, Russ Altman, Simran Arora, Sydney von Arx, Michael~S Bernstein, Jeannette Bohg, Antoine Bosselut, Emma Brunskill, et~al.
\newblock On the opportunities and risks of foundation models.
\newblock {\em arXiv preprint arXiv:2108.07258}, 2021.

\bibitem{chen2011msvd}
David~L. Chen and William~B. Dolan.
\newblock Collecting highly parallel data for paraphrase evaluation.
\newblock In {\em Proceedings of the 49th Annual Meeting of the Association for Computational Linguistics (ACL-2011)}, Portland, OR, June 2011.

\bibitem{chen2023video}
Jun Chen, Deyao Zhu, Kilichbek Haydarov, Xiang Li, and Mohamed Elhoseiny.
\newblock Video chatcaptioner: Towards enriched spatiotemporal descriptions.
\newblock {\em arXiv preprint arXiv:2304.04227}, 2023.

\bibitem{chen2021towards}
Shaoxiang Chen and Yu-Gang Jiang.
\newblock Towards bridging event captioner and sentence localizer for weakly supervised dense event captioning.
\newblock In {\em Proceedings of the IEEE/CVF Conference on Computer Vision and Pattern Recognition}, pages 8425--8435, 2021.

\bibitem{de2023socratic}
Irene de Zarz{\`a}, Joachim de Curt{\`o}, and Carlos~T Calafate.
\newblock Socratic video understanding on unmanned aerial vehicles.
\newblock {\em Procedia Computer Science}, 225:144--154, 2023.

\bibitem{epstein2019oops}
Dave Epstein, Boyuan Chen, and Carl Vondrick.
\newblock Oops! predicting unintentional action in video, 2019.

\bibitem{eysenbach2023role}
Gunther Eysenbach et~al.
\newblock The role of chatgpt, generative language models, and artificial intelligence in medical education: a conversation with chatgpt and a call for papers.
\newblock {\em JMIR Medical Education}, 9(1):e46885, 2023.

\bibitem{caba2015activitynet}
Bernard~Ghanem Fabian Caba~Heilbron, Victor~Escorcia and Juan~Carlos Niebles.
\newblock Activitynet: A large-scale video benchmark for human activity understanding.
\newblock In {\em Proceedings of the IEEE Conference on Computer Vision and Pattern Recognition}, pages 961--970, 2015.

\bibitem{feng2023llm4vg}
Wei Feng, Xin Wang, Hong Chen, Zeyang Zhang, Zihan Song, Yuwei Zhou, and Wenwu Zhu.
\newblock Llm4vg: Large language models evaluation for video grounding.
\newblock 2023.

\bibitem{Finkel2024BringingST}
Justin Finkel and Paul~A. O'Gorman.
\newblock Bringing statistics to storylines: rare event sampling for sudden, transient extreme events.
\newblock 2024.

\bibitem{floridi2020gpt}
Luciano Floridi and Massimo Chiriatti.
\newblock Gpt-3: Its nature, scope, limits, and consequences.
\newblock {\em Minds and Machines}, 30:681--694, 2020.

\bibitem{Gao2017TALLTA}
J. Gao, Chen Sun, Zhenheng Yang, and Ramakant Nevatia.
\newblock Tall: Temporal activity localization via language query.
\newblock {\em 2017 IEEE International Conference on Computer Vision (ICCV)}, pages 5277--5285, 2017.

\bibitem{gao2021fast}
Junyu Gao and Changsheng Xu.
\newblock Fast video moment retrieval.
\newblock In {\em Proceedings of the IEEE/CVF International Conference on Computer Vision}, pages 1523--1532, 2021.

\bibitem{Hao2023Fine}
Jiachang Hao, Haifeng Sun, Pengfei Ren, Yiming Zhong, Jingyu Wang, Qi Qi, and Jianxin Liao.
\newblock Fine-grained text-to-video temporal grounding from coarse boundary.
\newblock {\em ACM Trans. Multimedia Comput. Commun. Appl.}, 19(5), mar 2023.

\bibitem{he2019read}
Dongliang He, Xiang Zhao, Jizhou Huang, Fu Li, Xiao Liu, and Shilei Wen.
\newblock Read, watch, and move: Reinforcement learning for temporally grounding natural language descriptions in videos.
\newblock In {\em Proceedings of the AAAI Conference on Artificial Intelligence}, volume~33, pages 8393--8400, 2019.

\bibitem{lisa2017moment}
Lisa Hendricks, Oliver Wang, Eli Shechtman, Josef Sivic, Trevor Darrell, and Bryan Russell.
\newblock Localizing moments in video with natural language.
\newblock pages 5804--5813, 10 2017.

\bibitem{hendricks18didemo}
Lisa~Anne Hendricks, Oliver Wang, Eli Shechtman, Josef Sivic, Trevor Darrell, and Bryan Russell.
\newblock Localizing moments in video with temporal language.
\newblock In {\em Empirical Methods in Natural Language Processing (EMNLP)}, 2018.

\bibitem{jiao23survey_vad}
Runyu Jiao, Yi Wan, Fabio Poiesi, and Yiming Wang.
\newblock Survey on video anomaly detection in dynamic scenes with moving cameras.
\newblock {\em Artif. Intell. Rev.}, 56(Suppl 3):3515–3570, oct 2023.

\bibitem{jin2023emsassistdemo}
Liuyi Jin, Tian Liu, Amran Haroon, Radu Stoleru, Michael Middleton, Ziwei Zhu, and Theodora Chaspari.
\newblock Demo: Emsassist – an end-to-end mobile voice assistant at the edge for emergency medical services.
\newblock In {\em Proceedings of the 21st Annual International Conference on Mobile Systems, Applications and Services}, 2023.

\bibitem{jin2023emsassist}
Liuyi Jin, Tian Liu, Amran Haroon, Radu Stoleru, Michael Middleton, Ziwei Zhu, and Theodora Chaspari.
\newblock Emsassist: An end-to-end mobile voice assistant at the edge for emergency medical services.
\newblock In {\em Proceedings of the 21st Annual International Conference on Mobile Systems, Applications and Services}, pages 275--288, 2023.

\bibitem{kojima2022large}
Takeshi Kojima, Shixiang~Shane Gu, Machel Reid, Yutaka Matsuo, and Yusuke Iwasawa.
\newblock Large language models are zero-shot reasoners.
\newblock {\em Advances in neural information processing systems}, 35:22199--22213, 2022.

\bibitem{krishna2017dense}
Ranjay Krishna, Kenji Hata, Frederic Ren, Li Fei-Fei, and Juan Carlos~Niebles.
\newblock Dense-captioning events in videos.
\newblock In {\em Proceedings of the IEEE international conference on computer vision}, pages 706--715, 2017.

\bibitem{lan2023survey}
Xiaohan Lan, Yitian Yuan, Xin Wang, Zhi Wang, and Wenwu Zhu.
\newblock A survey on temporal sentence grounding in videos.
\newblock {\em ACM Transactions on Multimedia Computing, Communications and Applications}, 19(2):1--33, 2023.

\bibitem{li2024multimodal}
Chunyuan Li, Zhe Gan, Zhengyuan Yang, Jianwei Yang, Linjie Li, Lijuan Wang, Jianfeng Gao, et~al.
\newblock Multimodal foundation models: From specialists to general-purpose assistants.
\newblock {\em Foundations and Trends{\textregistered} in Computer Graphics and Vision}, 16(1-2):1--214, 2024.

\bibitem{li2023blip2}
Junnan Li, Dongxu Li, Silvio Savarese, and Steven Hoi.
\newblock Blip-2: Bootstrapping language-image pre-training with frozen image encoders and large language models.
\newblock In {\em International conference on machine learning}, pages 19730--19742. PMLR, 2023.

\bibitem{2023videochat}
Kunchang Li, Yinan He, Yi Wang, Yizhuo Li, Wenhai Wang, Ping Luo, Yali Wang, Limin Wang, and Yu Qiao.
\newblock Videochat: Chat-centric video understanding.
\newblock {\em arXiv preprint arXiv:2305.06355}, 2023.

\bibitem{li2023mvbench}
Kunchang Li, Yali Wang, Yinan He, Yizhuo Li, Yi Wang, Yi Liu, Zun Wang, Jilan Xu, Guo Chen, Ping Luo, Limin Wang, and Yu Qiao.
\newblock Mvbench: A comprehensive multi-modal video understanding benchmark, 2023.

\bibitem{liu2024visual}
Haotian Liu, Chunyuan Li, Qingyang Wu, and Yong~Jae Lee.
\newblock Visual instruction tuning.
\newblock {\em Advances in neural information processing systems}, 36, 2024.

\bibitem{liu2024eric}
Tian Liu, Liuyi Jin, Radu Stoleru, Amran Haroon, Charles Swanson, and Kexin Feng.
\newblock Eric: Estimating rainfall with commodity doorbell camera for precision residential irrigation.
\newblock {\em arXiv preprint arXiv:2409.13104}, 2024.

\bibitem{liu2024few}
Tian Liu, Huixin Zhang, Shubham Parashar, and Shu Kong.
\newblock Few-shot recognition via stage-wise augmented finetuning.
\newblock {\em arXiv preprint arXiv:2406.11148}, 2024.

\bibitem{luo2023valley}
Ruipu Luo, Ziwang Zhao, Min Yang, Junwei Dong, Da Li, Pengcheng Lu, Tao Wang, Linmei Hu, Minghui Qiu, and Zhongyu Wei.
\newblock Valley: Video assistant with large language model enhanced ability.
\newblock {\em arXiv preprint arXiv:2306.07207}, 2023.

\bibitem{Maaz2023VideoChatGPT}
Muhammad Maaz, Hanoona Rasheed, Salman Khan, and Fahad~Shahbaz Khan.
\newblock Video-chatgpt: Towards detailed video understanding via large vision and language models.
\newblock {\em arXiv:2306.05424}, 2023.

\bibitem{McMullan2023fireball}
Sarah McMullan, Denis Vida, Hadrien Devillepoix, Jim Rowe, Luke Daly, A. King, Martin Cupak, Robert Howie, Eleanor Sansom, Patrick Shober, Martin Towner, Seamus Anderson, Luke McFadden, Jana Horak, Andrew Smedley, Katherine Joy, Alan Shuttleworth, Francois Colas, Brigitte Zanda, and Gareth Collins.
\newblock The winchcombe fireball—that lucky survivor.
\newblock {\em Meteoritics \& Planetary Science}, 05 2023.

\bibitem{meta2024introducing}
AI Meta.
\newblock Introducing meta llama 3: The most capable openly available llm to date.
\newblock {\em Meta AI.}, 2024.

\bibitem{Miech2019HowTo100MLA}
Antoine Miech, Dimitri Zhukov, Jean-Baptiste Alayrac, Makarand Tapaswi, Ivan Laptev, and Josef Sivic.
\newblock Howto100m: Learning a text-video embedding by watching hundred million narrated video clips.
\newblock {\em 2019 IEEE/CVF International Conference on Computer Vision (ICCV)}, pages 2630--2640, 2019.

\bibitem{min2023recent}
Bonan Min, Hayley Ross, Elior Sulem, Amir Pouran~Ben Veyseh, Thien~Huu Nguyen, Oscar Sainz, Eneko Agirre, Ilana Heintz, and Dan Roth.
\newblock Recent advances in natural language processing via large pre-trained language models: A survey.
\newblock {\em ACM Computing Surveys}, 56(2):1--40, 2023.

\bibitem{parashar2024neglected}
Shubham Parashar, Zhiqiu Lin, Tian Liu, Xiangjue Dong, Yanan Li, Deva Ramanan, James Caverlee, and Shu Kong.
\newblock The neglected tails of vision-language models.
\newblock In {\em {IEEE/CVF} Conference on Computer Vision and Pattern Recognition (CVPR)}, 2024.

\bibitem{pramanick2023egovlpv2}
Shraman Pramanick, Yale Song, Sayan Nag, Kevin~Qinghong Lin, Hardik Shah, Mike~Zheng Shou, Rama Chellappa, and Pengchuan Zhang.
\newblock Egovlpv2: Egocentric video-language pre-training with fusion in the backbone.
\newblock In {\em Proceedings of the IEEE/CVF International Conference on Computer Vision}, pages 5285--5297, 2023.

\bibitem{radford2021learning}
Alec Radford, Jong~Wook Kim, Chris Hallacy, Aditya Ramesh, Gabriel Goh, Sandhini Agarwal, Girish Sastry, Amanda Askell, Pamela Mishkin, Jack Clark, et~al.
\newblock Learning transferable visual models from natural language supervision.
\newblock In {\em International conference on machine learning}, pages 8748--8763. PMLR, 2021.

\bibitem{radford2018improving}
Alec Radford, Karthik Narasimhan, Tim Salimans, Ilya Sutskever, et~al.
\newblock Improving language understanding by generative pre-training.
\newblock 2018.

\bibitem{rajagopalan2013self}
Shyam Rajagopalan, Abhinav Dhall, and Roland Goecke.
\newblock Self-stimulatory behaviours in the wild for autism diagnosis.
\newblock In {\em Proceedings of the IEEE International Conference on Computer Vision Workshops}, pages 755--761, 2013.

\bibitem{regneri2013grounding}
Michaela Regneri, Marcus Rohrbach, Dominikus Wetzel, Stefan Thater, Bernt Schiele, and Manfred Pinkal.
\newblock Grounding action descriptions in videos.
\newblock {\em Transactions of the Association for Computational Linguistics}, 1:25--36, 2013.

\bibitem{Sig2016Charades}
G.A. Sigurdsson, G. Varol, X. Wang, A. Farhadi, I. Laptev, and A. Gupta.
\newblock Hollywood in homes: Crowdsourcing data collection for activity understanding.
\newblock {\em Computer Vision – ECCV 2016. ECCV 2016. Lecture Notes in Computer Science(), vol 9905. Springer, Cham. https://doi.org/10.1007/978-3-319-46448-0\_31}, 2016.

\bibitem{tang2023video}
Yunlong Tang, Jing Bi, Siting Xu, Luchuan Song, Susan Liang, Teng Wang, Daoan Zhang, Jie An, Jingyang Lin, Rongyi Zhu, et~al.
\newblock Video understanding with large language models: A survey.
\newblock {\em arXiv preprint arXiv:2312.17432}, 2023.

\bibitem{tang2022multi}
Yunlong Tang, Siting Xu, Teng Wang, Qin Lin, Qinglin Lu, and Feng Zheng.
\newblock Multi-modal segment assemblage network for ad video editing with importance-coherence reward.
\newblock In {\em Proceedings of the Asian Conference on Computer Vision}, pages 3519--3535, 2022.

\bibitem{van2018species}
Grant Van~Horn, Oisin Mac~Aodha, Yang Song, Yin Cui, Chen Sun, Alex Shepard, Hartwig Adam, Pietro Perona, and Serge Belongie.
\newblock The inaturalist species classification and detection dataset.
\newblock In {\em 2018 IEEE/CVF Conference on Computer Vision and Pattern Recognition}, pages 8769--8778, 2018.

\bibitem{wang2022video}
Xin Wang, Xiaohan Lan, and Wenwu Zhu.
\newblock Video grounding and its generalization.
\newblock In {\em Proceedings of the 30th ACM International Conference on Multimedia}, pages 7377--7379, 2022.

\bibitem{wang2024hawkeye}
Yueqian Wang, Xiaojun Meng, Jianxin Liang, Yuxuan Wang, Qun Liu, and Dongyan Zhao.
\newblock Hawkeye: Training video-text llms for grounding text in videos.
\newblock {\em arXiv preprint arXiv:2403.10228}, 2024.

\bibitem{wu2020tree}
Jie Wu, Guanbin Li, Si Liu, and Liang Lin.
\newblock Tree-structured policy based progressive reinforcement learning for temporally language grounding in video.
\newblock In {\em Proceedings of the AAAI Conference on Artificial Intelligence}, volume~34, pages 12386--12393, 2020.

\bibitem{xie2023funqa}
Binzhu Xie, Sicheng Zhang, Zitang Zhou, Bo Li, Yuanhan Zhang, Jack Hessel, Jingkang Yang, and Ziwei Liu.
\newblock Funqa: Towards surprising video comprehension.
\newblock {\em arXiv preprint arXiv:2306.14899}, 2023.

\bibitem{Xu2016MSRVTT}
Jun Xu, Tao Mei, Ting Yao, and Yong Rui.
\newblock Msr-vtt: A large video description dataset for bridging video and language.
\newblock {\em 2016 IEEE Conference on Computer Vision and Pattern Recognition (CVPR)}, pages 5288--5296, 2016.

\bibitem{yang2023vid2seq}
Antoine Yang, Arsha Nagrani, Paul~Hongsuck Seo, Antoine Miech, Jordi Pont-Tuset, Ivan Laptev, Josef Sivic, and Cordelia Schmid.
\newblock Vid2seq: Large-scale pretraining of a visual language model for dense video captioning.
\newblock In {\em Proceedings of the IEEE/CVF Conference on Computer Vision and Pattern Recognition}, pages 10714--10726, 2023.

\bibitem{yuan2021closer}
Yitian Yuan, Xiaohan Lan, Xin Wang, Long Chen, Zhi Wang, and Wenwu Zhu.
\newblock A closer look at temporal sentence grounding in videos: Dataset and metric.
\newblock In {\em Proceedings of the 2nd international workshop on human-centric multimedia analysis}, pages 13--21, 2021.

\bibitem{zahan2023human_gesture_autism}
S. Zahan, Z. Gilani, G. Hassan, and A. Mian.
\newblock Human gesture and gait analysis for autism detection.
\newblock In {\em 2023 IEEE/CVF Conference on Computer Vision and Pattern Recognition Workshops (CVPRW)}, pages 3328--3337, Los Alamitos, CA, USA, jun 2023. IEEE Computer Society.

\bibitem{zhang-etal-2023-videollama}
Hang Zhang, Xin Li, and Lidong Bing.
\newblock Video-{LL}a{MA}: An instruction-tuned audio-visual language model for video understanding.
\newblock In Yansong Feng and Els Lefever, editors, {\em Proceedings of the 2023 Conference on Empirical Methods in Natural Language Processing: System Demonstrations}, pages 543--553, Singapore, Dec. 2023. Association for Computational Linguistics.

\bibitem{zhang2020span}
Hao Zhang, Aixin Sun, Wei Jing, and Joey~Tianyi Zhou.
\newblock Span-based localizing network for natural language video localization.
\newblock {\em arXiv preprint arXiv:2004.13931}, 2020.

\bibitem{zhang2023temporal}
Hao Zhang, Aixin Sun, Wei Jing, and Joey~Tianyi Zhou.
\newblock Temporal sentence grounding in videos: A survey and future directions.
\newblock {\em IEEE Transactions on Pattern Analysis and Machine Intelligence}, 45(8):10443--10465, 2023.

\bibitem{Zhang2023TextVisPrompt}
Y. Zhang, X. Chen, J. Jia, S. Liu, and K. Ding.
\newblock Text-visual prompting for efficient 2d temporal video grounding.
\newblock In {\em 2023 IEEE/CVF Conference on Computer Vision and Pattern Recognition (CVPR)}, pages 14794--14804, Los Alamitos, CA, USA, jun 2023. IEEE Computer Society.

\bibitem{zhao2023learning}
Yue Zhao, Ishan Misra, Philipp Kr{\"a}henb{\"u}hl, and Rohit Girdhar.
\newblock Learning video representations from large language models.
\newblock In {\em Proceedings of the IEEE/CVF Conference on Computer Vision and Pattern Recognition}, pages 6586--6597, 2023.

\bibitem{zheng2024training}
Minghang Zheng, Xinhao Cai, Qingchao Chen, Yuxin Peng, and Yang Liu.
\newblock Training-free video temporal grounding using large-scale pre-trained models.
\newblock {\em arXiv preprint arXiv:2408.16219}, 2024.

\end{thebibliography}
}

\newpage
\clearpage

\section*{Appendix}
\subsection*{A. Prompts Used}
We list the prompts used for the Vid-LLM and VLM-LLM approaches on this paper in \autoref{tab:prompt design}.
\label{sec: appendix}

\begin{table}[h]

  \centering
  \caption{Prompts used for Vid-LLMs and VLM-LLM approaches.}
  \footnotesize{
  \scalebox{1}{
  \begin{tabular}{c|p{1.5cm}|p{4cm}}
    \toprule
    \centering Methods & \centering Prompt Types & \centering Prompt Content \tabularnewline [0.5ex]
    \midrule
     \multirow{5}{*}{\makecell{Vid-\\LLM}} & \centering \multirow{5}{*}{prompt} &  "Find the start time and end time of the query below from the video. \newline 
    Query: [Annotated Activity Description]" \tabularnewline [0.5ex]
    \midrule
    \midrule
    \multirow{22}{*}{\makecell{VLM-\\LLM}} & \centering \multirow{6}{*}{\begin{tabular}[c]{@{}c@{}}System\\instruction\end{tabular}} & "You are a video analyst. You can read the video text representation and infer the start and end time of a given activity from the cue words found in the video text representation." \tabularnewline [0.5ex]
    \cmidrule{2-3}
    & \centering \multirow{13}{*}{Query} & "Find the start time and end time of the query below given the video text representation. Even if the query is not present in the description, try to find relationship between the meaning of words and infer. You must predict an answer and do not predict any null prediction. Give your answer in json format. \newline
    Query: [Annotated Activity Description]" \tabularnewline [0.5ex]
    \cmidrule{2-3}
    & \centering \centering \multirow{3}{*}{\begin{tabular}[c]{@{}c@{}}Localization\\Prompt\end{tabular}} & [Query] \newline
    Video Text Representation: [$Text\_Rep(V)$]\tabularnewline [0.5ex]
    \bottomrule
  \end{tabular}
  }}
  \label{tab:prompt design}
  \vspace{-5mm}
\end{table}

\end{document}